\documentclass[runningheads,a4paper]{llncs}

\usepackage{graphicx}
\usepackage[caption=false]{subfig}
\usepackage{url}

\makeatletter
\newcommand{\@chapapp}{\relax}%
\makeatother
\usepackage[title,toc,titletoc,header]{appendix}

\usepackage{algorithm}
\usepackage[noend]{algpseudocode}
\newcommand{\Break}{\State \textbf{break} }
\usepackage{tablefootnote}

\usepackage{multirow}

\usepackage{makecell}

\usepackage{amsmath,amssymb,amstext}
\usepackage{float}

\begin{document}

\mainmatter  

\title{Theano-MPI: a Theano-based Distributed Training Framework}
\titlerunning{Theano-MPI: a Theano-based Distributed Training Framework}

\author{He Ma \inst{1} \and Fei Mao \inst{2} \and Graham W.~Taylor \inst{1}}

\institute{School of Engineering, University of Guelph, CA \email{\{hma02,gwtaylor\}@uoguelph.ca} \and SHARCNET, Compute Canada, CA \email{feimao@sharcnet.ca} }

\maketitle

\begin{abstract} 
	
We develop a scalable and extendable training framework that
can utilize GPUs across nodes in a
cluster and accelerate the training of deep learning
models based on data parallelism. Both synchronous
and asynchronous training are implemented in our
framework, where parameter exchange among
GPUs is based on CUDA-aware MPI. In
this report, we analyze the convergence
and capability of the framework to reduce training time when
scaling the synchronous training of AlexNet and
GoogLeNet from 2 GPUs to 8 GPUs. In addition, we explore
novel ways to reduce the communication overhead caused
by exchanging parameters. Finally, we release the framework as
open-source for further research on distributed deep learning\footnote{\url{https://github.com/uoguelph-mlrg/Theano-MPI}}.
	
\end{abstract} 

\section{Introduction}

With the constant improvement of hardware and discovery of new
architectures, algorithms, and applications, deep learning is gaining
popularity in both academia and industry. Object recognition
\cite{Russakovsky15}, is now dominated by deep learning methods, which
in many cases, rival human performance. Recent success in areas such
as activity recognition from video \cite{Karpathy14} and statistical
machine translation \cite{Koehn12} is an example of deep learning's
ascent both in performance and at scale.

With the new generations of GPU cards and
increased device memory, researchers are able to design and train
models with more than 140 million parameters (c.f.~VGGNet
\cite{Simonyan14}) and models that are as deep as 150 layers
(c.f.~ResNet \cite{He15}).

The emergence of larger datasets, e.g.~ImageNet
\cite{Russakovsky15} and MS-COCO \cite{Lin14COCO}, 
challenges artificial intelligence research
and leads us to design deeper and more expressive models so
that the complexity of models is sufficient for the task.

Despite of the increased computing power of GPUs, 
it usually takes weeks to train such large models
to desired accuracy on a single GPU. This is due to the
increased time associated with training deeper models and 
iterating over the examples in larger datasets. This is where
distributed training of deep learning models becomes crucial, especially
for activities such as model search which may involve training
and evaluating models thousands of times.

A na\"ive approach to scaling up is running several copies of the
same model in parallel on multiple computing resources (e.g.~GPUs),
each computing its share of the dataset and averaging
their parameters at every iteration. This strategy is called
data parallelism, and its efficient implementation is the focus
of our work. More sophisticated forms of distributed training,
including model parallelism are important but outside the
current scope of our framework.

Theano \cite{2016arXiv160502688short} is an open-source Python library for
developing complex algorithms via mathematical 
expressions. It is often used for
facilitating machine learning research. Its support
for automatic symbolic differentiation and
GPU-accelerated computing has made it popular
within the deep learning community.
Like other deep learning
platforms, including Cafe \cite{Jia14}, Torch
\cite{Collobert11}, TensorFlow \cite{Abadi16} and MXNet
\cite{Chen15}, Theano uses CUDA as one of its main
backends for GPU accelerated computation. Since a
single GPU is limited by its device memory and available threads
when solving compute-intensive
problems, very recently researchers have started to build 
multi-GPU support into the most popular frameworks. This includes
the multi-GPU version of Caffe (FireCaffe \cite{Iandola15}),
Torch and Theano (Platoon).

Because the Theano environment usually compiles models for one GPU per process, 
we need to drive multiple GPUs using multiple processes. So finding a way to 
communicate between processes becomes a fundamental problem within a multi-GPU framework. 
There are several existing approaches of implementing inter-process 
communication besides manually programming on sockets, such as Signals, 
Message Queues, Message Passing, Pipes, Shared Memory,
Memory Mapped Files, etc. However, among those approaches, Message 
Passing is most suitable for collective communication between multiple 
programs across a cluster because of its well-developed point-to-point and 
collective protocols. Message Passing Interface (MPI) is a language-independent 
communication protocol that can undertake the task of inter-process 
communication across machines. It is a standardized message-passing 
system designed for programming on large-scale parallel applications.

Parameter transfer is a basic operation in the distributed training of
deep learning models.
Therefore, the transfer speed between processes severely impacts the
overall data throughput speedup\footnote{We define data throughput
  speedup as the change in total time taken to process a certain
  amount of examples. It includes both training and communication time.}. Since the parameters to be transferred are 
computed on GPUs, a GPU-to-GPU transfer is required. 
Compared to the basic \verb|transfer()| function in Theano, NVIDIA GPUDirect P2P technology 
makes this possible by transferring data between GPUs without passing 
through host memory. Specifically, it enables CUDA devices to perform direct read 
and write operations on other CUDA host and device memory. In the context 
of MPI, GPUDirect P2P technology allows a \verb|GPUArray| memory buffer to be 
transferred in basic point-to-point and collective operations, making MPI ``CUDA-Aware''.

Leveraging CUDA-aware MPI, we have developed a scalable training
framework that provides multi-node and multi-GPU support to Theano and
efficient inter-GPU parameter transfer at the same time. To the best
of our knowledge, this is to-date the most convenient way to deploy
Theano processes on a multi-node multi-GPU cluster.

\section{Related Work}

The idea of exploiting data parallelism in machine
learning has been widely explored in recent years in both
asynchronous and synchronous ways. To accelerate
the training of a speech recognition model on distributed
CPU cores, DownPour, an asynchronous parameter exchanging method
\cite{Dean12}, was proposed. It was the largest-scale
method to-date for distributed training of neural networks.
It was later found that controlling the
maximum staleness of parameter updates received by the
server leads to faster training convergence \cite{Ho13} on
problems like topic modeling, matrix factorization and lasso
regression compared to a purely asynchronous approach. For 
accelerating image classification on the CIFAR and ImageNet 
datasets, an elastic averaging strategy between asynchronous
workers and the server was later proposed \cite{Zhang15}.
This algorithm allows more exploration of local optima than
DownPour and alleviates the need for frequent communication 
between workers and the server.

Krizhevsky proposed his trick on parallelizing
the training of AlexNet \cite{Krizhevsky12} on multiple GPUs
in a synchronous way \cite{Krizhevsky14}. This work showed that
eight GPU workers training on the same batch size of 128 can
give up to 6.25$\times$ data throughput speedup and nearly the same
convergence as trained on a single GPU when exploiting both model and
data parallelism. Notably, the increase in effective batch
size\footnote{effective batch size $=$ batch size $\times$ number of workers}
leads to very small changes in the final convergence of
AlexNet when the learning rate is scaled properly. Following his work,
a Theano-based two-GPU synchronous framework \cite{Ding14}
for accelerating the training of AlexNet was proposed, where both
weights and momentum are averaged between two GPUs
after each iteration. The model converges to the
same level as using a single GPU but in less time.

There has been more development on the acceleration of vision-based
deep learning in recent years. NVIDIA developed a multi-GPU deep learning
framework, DIGITS, which shows 3.5$\times$ data throughput speedup when
training AlexNet on 4 GPUs. Purine \cite{Lin14Purine} pipelines the propagation
of gradients between iterations and overlaps the communication
of large weights in fully connected layers with the rest of
back-propagation, giving near 12$\times$ data throughput speedup when
training GoogLeNet \cite{Szegedy14} on 12 GPUs.
Similarly, MXNet \cite{Chen15} also shows a super-linear data throughput
speedup on training GoogLeNet under a distributed training setting.

The Platoon project is a multi-GPU extension for Theano, created and 
maintained by the official Theano team. It currently supports only asynchronous 
data parallelism \emph{inside one compute node} based on \verb|posix_ipc| 
shared memory. In comparison, our framework, Theano-MPI, is designed to support 
GPUs that are distributed over multiple nodes in a cluster,  providing convenient process 
management and faster inter-GPU memory exchanging based on CUDA-aware MPI.
\section{Implementation}

Our goal is to make the field of distributed
deep learning more accessible by developing a scalable training
framework with two key components. First is Theano as a means of constructing an architecture
and optimizing it by Stochastic Gradient Descent (SGD).
Second is Massage Passing Interface (MPI) as an inter-process parameter
exchanger. We also aim to explore various ways to reduce communication
overhead in parallel SGD and expose some phenomena that affect
convergence and speedup when training deep learning models in a
distributed framework.

\subsection{The BSP Structure}

\begin{figure}[h]%
	  \vskip -0.1in
       \centering
       \subfloat[non-CUDA-aware\label{f:bsp_nca}]{\includegraphics[scale=0.5]{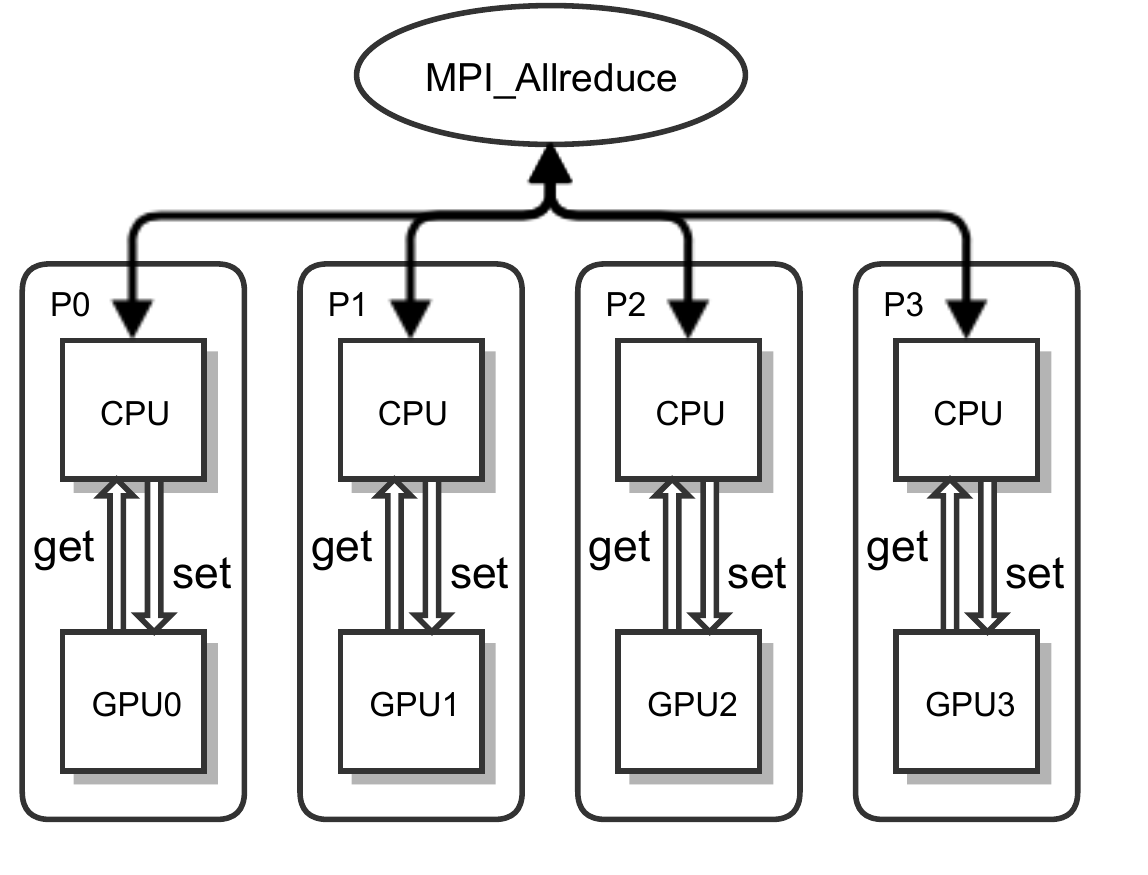}}\quad
       \subfloat[CUDA-aware\label{f:bsp_ca}]{\includegraphics[scale=0.5]{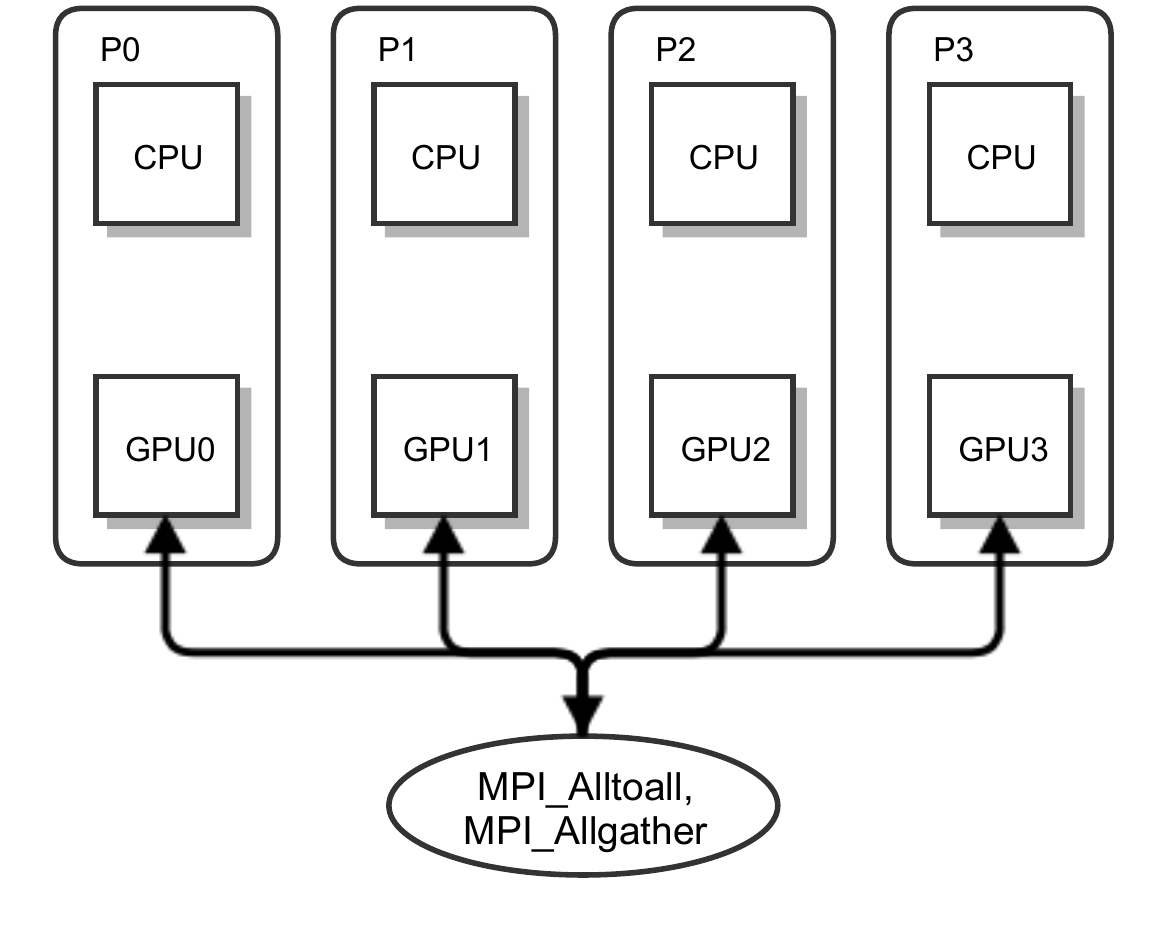}}
       \caption{A 4-GPU example of the BSP structure where arrows
         indicate communication for parameter exchange.}%
       \label{f:4GPUBSP}%
       \vskip -0.2in
\end{figure}

Bulk Synchronous Parallel (BSP) \cite{Leslie90} is
an intuitive way to implement parallel computing. In the BSP paradigm, workers proceed
with training in a synchronous way. Figure~\ref{f:bsp_nca}
shows a 4 GPU example of the proposed BSP structure
where the same model is built and run within four
processes, $P_0,P_1,P_2,P_3$. Each process uses one CPU and one
GPU. After the model's training graph is
compiled on the GPU, those parameters in the graph
become arrays in GPU memory whose values can be
retrieved from device to host and set from host to
device. When training starts, the training dataset
is split into four parts. In every iteration, each worker process
takes a mini-batch of examples from its share and performs
SGD on it. After that, all workers
are synchronized and model parameters are exchanged
between worker processes in a collective way.

\subsection{CUDA-aware Parameter Exchanging}

\begin{figure}[h]
	\vskip -0.1in
	\begin{center}
		\centerline{\includegraphics[scale=0.45]{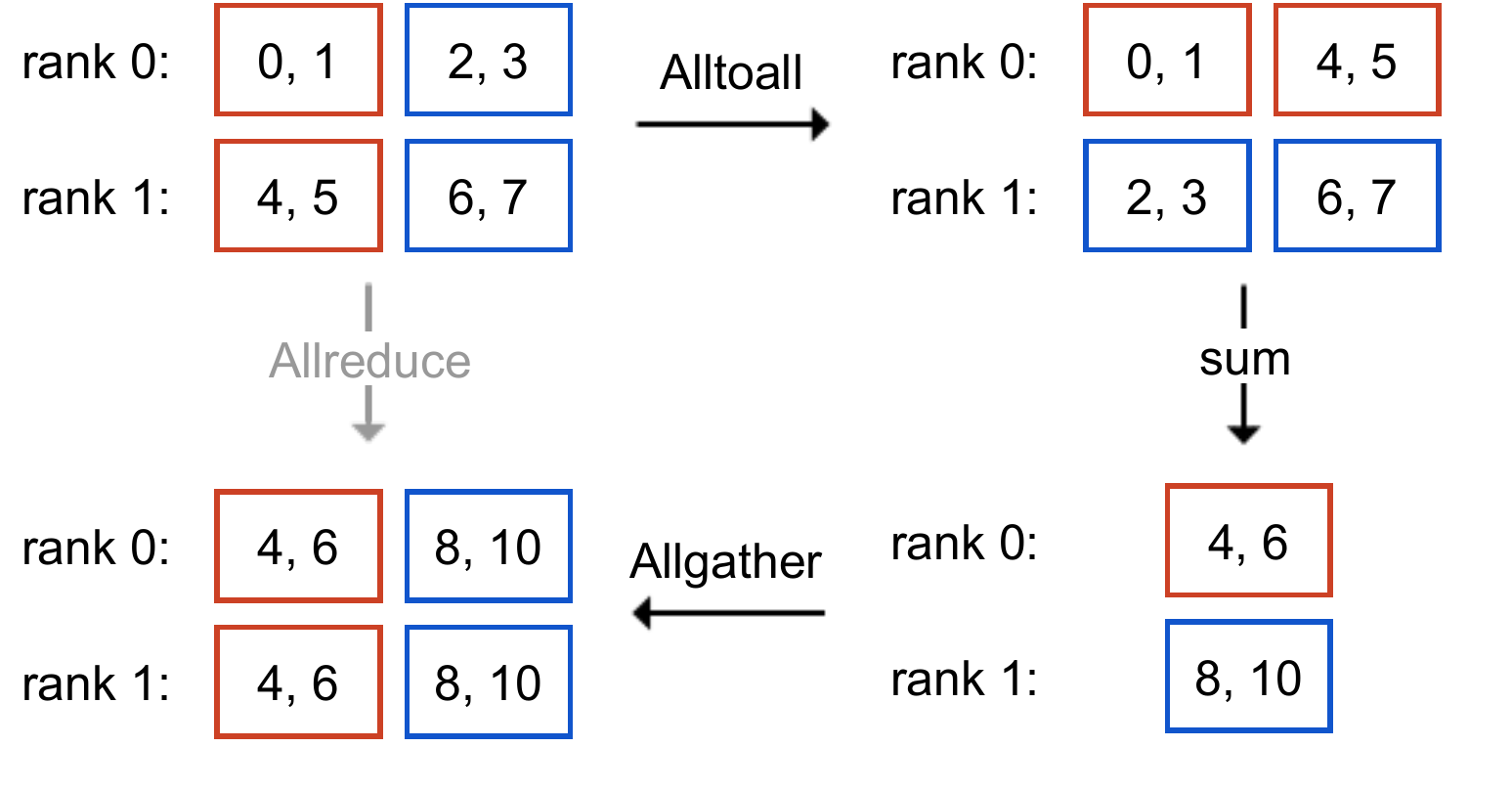}}
		\caption{An example demonstrating the reduction of arrays on rank 0 
			and rank 1 with the proposed Alltoall-sum-Allgather strategy compared 
			to MPI Allreduce. Sub-arrays of data items
                        (indicated by same-coloured boxes) need to be
                        summed and the results exchanged with the other ranks.}
		\label{f:asa-example}
	\end{center}
	\vskip -0.2in
\end{figure}

Synchronous parameter exchange is an array reduction problem which
consists of both data transfer and calculation. The GPUDirect P2P
technology allows exchanging parameters between GPUs without passing
through host memory, making MPI functions ``CUDA-aware''. Based on
this, we explored various strategies trying to minimize the data 
transfer and calculation time, and make more efficient use of QPI,
PCIe and network card bandwidth during data transfer. The basic
strategy is to use the MPI \verb|Allreduce()| function. However, the
CUDA-aware version of it in OpenMPI 1.8.7 does not give much
improvement since any collective MPI function with arithmetic
operations still needs to copy data to host memory. Functions like
\verb|Alltoall()| and \verb|Allgather()| do not involve any arithmetic
and therefore the CUDA-aware version of them (Fig.~\ref{f:bsp_ca}) can avoid passing through
host memory unless data transfer crossing the QPI bus is needed. We
therefore implemented a CUDA-aware \verb|Alltoall-sum-Allgather|
strategy which separates the data transfer and computation. An example
of this strategy is demonstrated in Fig.~\ref{f:asa-example}. Here,
the summation kernels required for parameter exchange are
executed in parallel on GPUs. Our test shows the 
GPU summation kernel takes only 1.6\% of the total communication time. 

Using low precision data types for weights or activations (or both) in
the forward pass during training of deep neural networks has received
much recent interest \cite{Courbariaux14,courbariaux2016binarynet}. It
was shown that training Maxout \cite{Goodfellow13} networks at 10 bits
fixed point precision can still yield near state-of-art test accuracy
\cite{Courbariaux14}.  In light of this, we also implemented the
transfer of parameters at half-precision while summing them at full
precision, in order to further reduce communication overhead.

\begin{figure}[h]
	\vskip -0.1in
	\begin{center}
		\centerline{\includegraphics[scale=0.62]{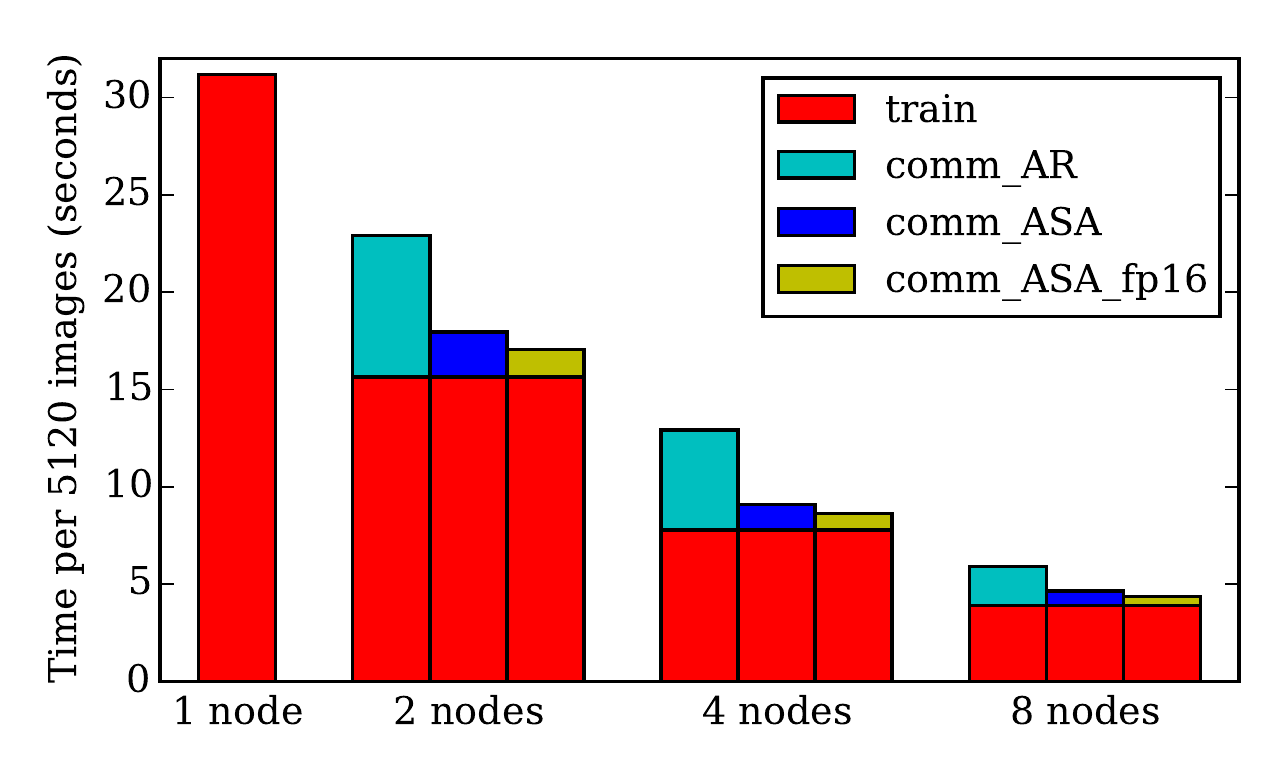}}
		\caption{Computation (train) vs.~ relative communication overhead of different parameter
			exchanging strategies during training AlexNet-128b
			(AR: Allreduce, ASA: CUDA-aware Alltoall-sum-Allgather).}
		\label{f:compare_time}
	\end{center}
	\vskip -0.2in
\end{figure}

Figure~\ref{f:compare_time} shows the improvement of the combination of strategies
over MPI \verb|Allreduce|. The ``ASA'' strategy shows three times
faster communication relative to MPI \verb|Allreduce| and the half
precision version of it gives nearly 6 times faster
performance. Those results are obtained on distributed GPUs on 8 nodes
in a cluster. Each node hosts one GPU.

Due to the limitation imposed by the Global Interpreter Lock (GIL) in
Python, overlapping the communication with the gradient calculation
as in \cite{Lin14Purine} has not yet been implemented in our
framework. We expect this, if implemented, would substantially reduce
the communication cost of exchanging large matrices in fully-connected
layers.
 
\subsection{Parallel Loading}

For large-scale visual recognition applications such as ImageNet LSVRC, the data required for training
is on the order of hundreds of Gigabytes. Therefore, it is difficult to
load all image data completely into memory after
training starts. Instead, images are stored as
batch files on local or remote disks and loaded one file
at a time by each process. Loading image batches \emph{x} from 
disk can be time consuming\footnote{Loading labels \emph{y}, on 
	the other hand, is much faster, therefore labels can be loaded completely into memory.}. 
It is affected by various factors, including file size, file format, disk I/O capability and network
 bandwidth if reading from remote disks. If in every iteration, 
the training process should wait for data loading to be ready in order to 
proceed, one can imagine the time cost by loading data will be 
critical to the total performance. One way to circumvent this, given the independence 
of loading and training, is to load those files in parallel with the forward and backward propagations
on the last loaded batch. However, this assumes loading one batch of images 
takes shorter than one iteration of training the model. This auxiliary 
loading process should follow procedures in Alg.~\ref{alg:paraload} to 
collaborate efficiently with its corresponding training process:

\begin{algorithm}[h]
	\caption{The parallel loading process}\label{alg:paraload}
	\begin{algorithmic}[1]
		\Require 
			\Statex Host memory allocated for loading image batch $hostdata_x$. 
			\Statex GPU memory allocated for preprocessed image batch $gpudata_x$ 
			\Statex GPU memory allocated for the actual model graph input $input_x$, 
			\Statex $mode$=None, $recv$=None, $filename$=None.
			\Statex Mean image $image\_mean$
		\Ensure
			\While{True}
			\State Receive the $mode$ (train, validate or stop) from training process
			 \If{$recv$=``stop''}
				 \Break
			\Else
				\State $mode \gets recv$
			\EndIf
			\State Receive the first filename to be loaded from training process $filename \gets recv$
			
			\While{True}
			\State Load file `` filename'' from disk into host memory $hostdata_x$.
			\State $hostdata_x$ = $hostdata_x$ $-$ $image\_mean$
			\State Crop and mirror $hostdata_x$ according to $mode$.
			\State Transfer $hostdata_x$ from host to GPU device memory $gpudata_x$.
			\State Wait for training on the last $input_x$ to finish by receiving the next filename to be loaded.
			\If{$recv$ in [``stop'', ``train'', ``val'']}
				\Break
			\Else
				\State $filename \gets recv$
			\EndIf 
				
			\State Transfer $gpudata_x$ to $input_x$.
			\State Synchronize GPU context.
			\State Notify training process to precede with the newly loaded $input_x$
			\EndWhile
			
			\EndWhile
	\end{algorithmic}
\end{algorithm}

 Different from the \verb|multiprocessing| and \verb|Queue| 
 messaging method in \cite{Ding14}, we used the MPI \verb|Spawn| 
 function to start a child process from each training process and 
 used the resulting MPI intra-communicator to pass messages 
 between the training process and its child process. As shown in 
 Algorithm \ref{alg:paraload}, the parallel loading process can 
 read image files, subtract the mean image, crop sub-images 
 and finally load preprocessed data onto GPUs. By doing this, 
 we are able to overlap the most compute-intensive part (Step 
 10 to 13 in Algorithm \ref{alg:paraload}) with forward and 
 backward graph propagation in the training process.

\section{Benchmarking}

Exchanging parameters is a necessary aspect of parallel SGD, however,
it can be achieved in a variety of different ways. Parameters updated
during SGD include weights (and biases), momentum (if using momentum 
SGD) and raw gradients. 
Averaging \emph{weights} after gradient descent (AWAGD) \cite{Krizhevsky14,Ding14}
is a straightforward parallel SGD scheme. We have
proved \cite{Ma15} that training a perceptron using this
scheme on multiple GPUs can either be equivalent to or a close
approximate of sequential SGD training on a single GPU depending on whether
or not effective batch size is kept constant. In this scheme, the
learning rate is scaled with the number of GPUs used \cite{Krizhevsky14},
namely $k$. It can also be shown that this scheme is equivalent
to summing up the \emph{parameter updates} from all GPUs
before performing gradient descent (SUBGD), which does not require
scaling up the learning rate. However, our experiments show that tuning
the learning rate is still dependent on $k$
to ensure initial convergence of the model.
Table~\ref{t:lr} lists the learning rates we used and the convergence we achieved in training
AlexNet\footnote{Top-5 error at epoch 62. The implementation is based
	on theano\_alexnet from uoguelph-mlrg.
	\url{https://github.com/uoguelph-mlrg/theano_alexnet}.}
and GoogLeNet\footnote{Top-5 error at epoch 70. BVLC GoogLeNet
	implementation in Caffe is referenced in building the model.
	\url{https://github.com/BVLC/caffe/tree/master/models/bvlc_googlenet}.
	* At the time of submission, the GoogLeNet 1GPU test is still ongoing.
	The top-5 error is taken from \cite{Szegedy14}.}
at different scales (number of workers).

Recent work has applied low precision to weights and activations
during training \cite{Courbariaux14}. In the extreme, binary data
types have been considered \cite{courbariaux2016binarynet}. This
enables efficient operation of the low-precision network both at
deployment (test time) and during the forward propagation stage during
training. However, gradients used for parameter updates must still be
stored at high-precision or learning will fail. Related to this, we
see a drop in accuracy due to reduced-precision parameter
exchange. The validation top-5 error of the ``8GPU-32b'' AlexNet in
Table 1 increased from 19.9\% to 20.3\% and that of the ``8GPU''
GoogLeNet increased from 10.65\% to 11.7\%.  Parameter exchange is an important
part of the update stage in a distributed training
framework. Therefore, there is a necessary accuracy-speed tradeoff to
consider when adopting a low-precision strategy.

\begin{table}[h]
  \caption{Trade-off between accuracy and speedup under different hyper parameter settings in training AlexNet
    and GoogLeNet based on the ASA strategy. The learning rate
    reported was the best one found empirically for the
    particular setting (HP: hyper-parameters, LR: learning
    rate, BS: batch size).}
	\label{t:lr}
	\begin{center}
		\begin{tabular}{c cccc|cccc}
			\hline\noalign{\smallskip}
			\multicolumn{1}{c}{\multirow{3}{*}{\makecell{\\ \# of \\ workers}}} & \multicolumn{4}{c|}{AlexNet} & \multicolumn{4}{c}{GoogLeNet} \\
			
			\noalign{\smallskip}
			\cline{2-9}
			\noalign{\smallskip}
			
			& \multicolumn{2}{c}{HP} & \multicolumn{2}{c|}{Result} &
			\multicolumn{2}{c}{HP} &  \multicolumn{2}{c}{Result} \\
			
			\noalign{\smallskip}
			\cline{2-9}
			\noalign{\smallskip}
			
			& LR & BS & Accuracy & Speedup & LR & BS & Accuracy & Speedup \\
			
			\noalign{\smallskip}
			\cline{1-9}
			\noalign{\smallskip}
			
			1GPU & 0.01 & 128 & 19.8\% & 1$\times$ & 0.01 & 32 & 10.07\%* & 1$\times$\\
			2GPU & 0.01 & 128 & 19.8\% & 1.7$\times$ & 0.007 & 32 & 10.20\% & 1.9$\times$\\
			4GPU & 0.01 &  128 & 20.4\% & 3.4$\times$ & 0.005 & 32 & 10.48\% & 3.7$\times$\\
			8GPU & 0.005 & 128 & 20.7\% & 6.7$\times$ & 0.005 & 32 & 10.65\% & 7.2$\times$\\
			8GPU & 0.005 & 32 & 19.9\% & 4.9$\times$ & \multicolumn{4}{c}{-} \\
			8GPU-fp16 & 0.005 & 32 & 20.3\% & 5.7$\times$ & 0.005 & 32 & 11.75\% & 7.3$\times$\\
			\noalign{\smallskip}
			\hline
		\end{tabular}
	\end{center}
	\vskip -0.2in
\end{table}

\begin{figure}[h]
	\begin{center}
		\centerline{\includegraphics[scale=0.7]{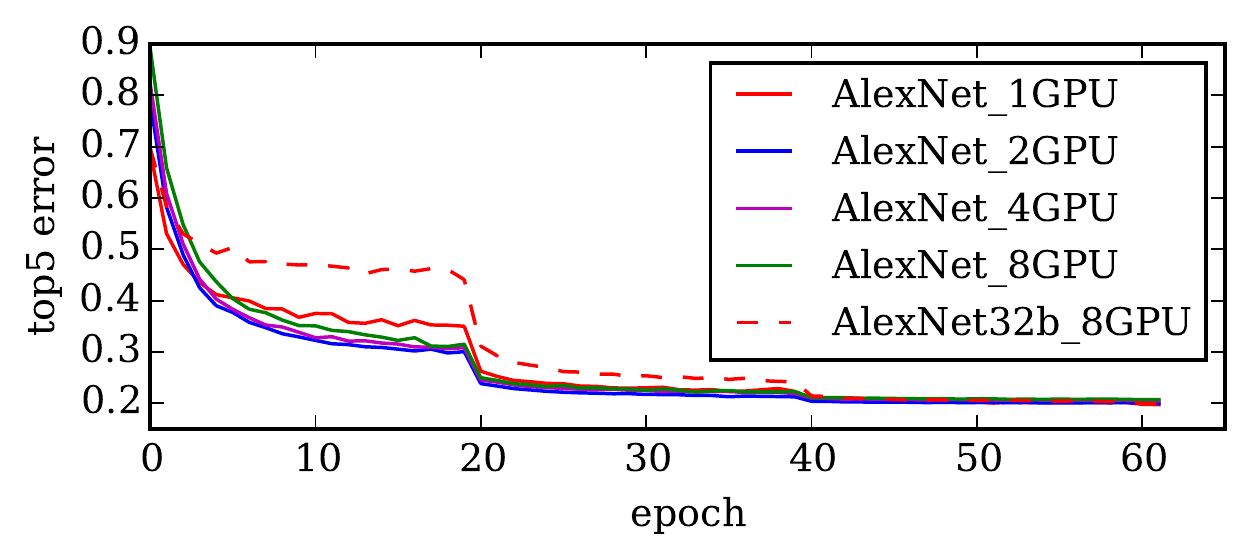}}
		\caption{Validation top-5 error of AlexNet trained
			at different scales (and batch sizes). Best
			viewed in colour.}
		\label{f:SUBGD_alexnet}
	\end{center}
	\vskip -0.2in
\end{figure}

\begin{figure}[h]
	\vskip -0.2in
	\begin{center}
		\centerline{\includegraphics[scale=0.7]{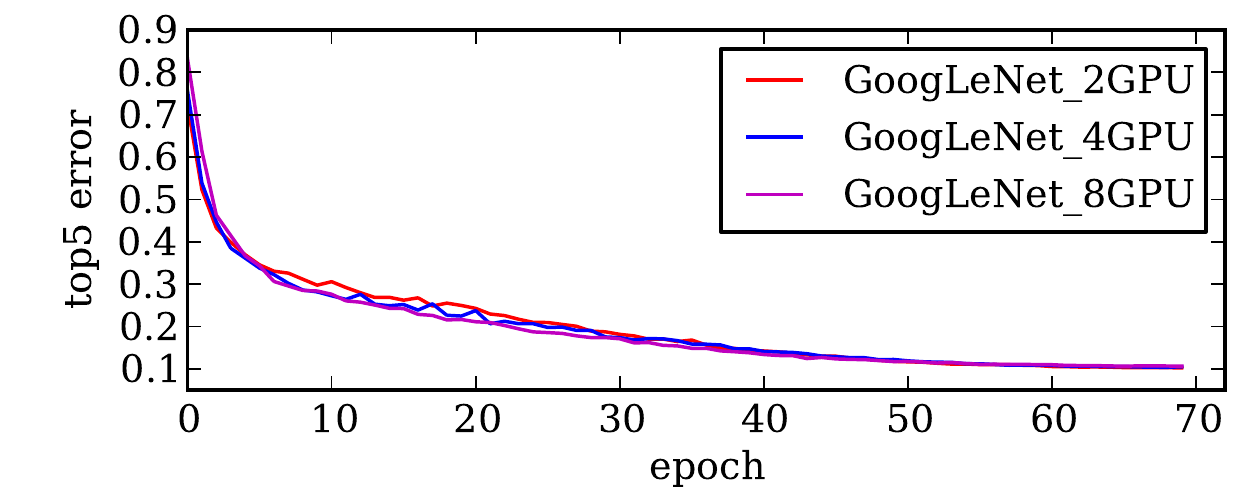}}
		\caption{Validation top-5 error of GoogLeNet trained
			at different scales. Best viewed in colour.}
		\label{f:SUBGD_googlenet}
	\end{center}
	\vskip -0.2in
\end{figure}

Figures~\ref{f:SUBGD_alexnet} and \ref{f:SUBGD_googlenet} show the
convergence of two models trained with SUBGD and the
\verb|Alltoall-sum-Allgather| strategy, in which AlexNet is trained on
1, 2, 4 and 8 GPUs with momentum SGD and 128 batch size on each GPU\footnote{Tested on the ILSVRC14 dataset
  \cite{Russakovsky15}.}. Similarly, GoogLeNet is trained on 2, 4 and
8 GPUs with a batch size of 32.  We see that as more workers are used,
the effective batch size becomes too large and the approximation from
parallel SGD to sequential SGD becomes worse.  As shown in Table
\ref{t:lr}, one way to preserve convergence at such a large-scale is to
reduce the batch size on each worker so that the effective batch size
stays small. This gives the model more potential to explore further at
low learning rates, though the accuracy improvement at the beginning
is slow. However, using smaller batch sizes means more frequent
parameter exchanges between workers, which demands attention toward
further reducing the communication overhead.

The speedup of training AlexNet and GoogLeNet are evaluated on 8
distributed GPU nodes (1 GPU per node). To show the performance of
accelerating larger models, we build VGGNet and test its scaling
performance on 8 GPUs in a single node with shared memory. This setup
meets the memory requirements of VGGNet.
Table~\ref{t:model_size} gives an overview of the structural
difference between those three models. Table~\ref{t:cross} reports the
training and communication time taken to process 5,120 images across
different models. We see that these three models scale differently in
the framework due to differences in the complexity of their operations
as well as the number of free parameters. CUDA-aware parameter
exchanging helps boost the speedup of the framework, especially when
the number of parameters is relatively large.

\begin{table}[h]
	\caption{Structural comparison between the three architectures
           which were implemented for benchmarking.}
	\label{t:model_size}
	\begin{center}
			\begin{tabular}{lcr}
                           \hline\noalign{\smallskip}
                          Model & Depth \tablefootnote{In terms
                                  of the amount of parameter-containing
                                  layers.}& \# of parameters\tablefootnote{In terms of the amount of float32 parameters. }\\
                          \noalign{\smallskip}
                          \hline
                          \noalign{\smallskip}
                          AlexNet&8&60,965,224\\
                          GoogLeNet&22&13,378,280\tablefootnote{This
                                        includes the parameters
                                        of the two auxiliary classifiers.}\\
                          VGGNet&19&138,357,544\\
                          \noalign{\smallskip}
                          \hline
			\end{tabular}
	\end{center}
	\vskip -0.2in
\end{table}

\begin{table}[h]
	\caption{Communication overhead per 5,120 images (s) / 
		speedup on 8 GPUs for different models
		(AR: Allreduce, ASA: CUDA-aware Alltoall-sum-Allgather, ASA16:
		CUDA-aware Alltoall-sum-Allgather
		w/ float16).}
	\label{t:cross}
	\begin{center}
			\begin{tabular}{lcccc}
				\hline\noalign{\smallskip}
				Model &Train(1GPU)&AR&ASA&ASA16\\
				\noalign{\smallskip}
				\hline
				\noalign{\smallskip}
				AlexNet-128b&3.90(31.2)&2.01/5.3$\times$&0.75/6.7$\times$&0.47/7.1$\times$\\
				AlexNet-32b&4.56(36.40)&8.03/2.9$\times$&2.94/4.9$\times$&1.83/5.7$\times$\\
				GoogLeNet-32b&16.82(134.9)&2.07/7.1$\times$&1.96/7.2$\times$&1.76/7.3$\times$\\
				VGGNet-32b&51.79(405.2)&41.41/4.3$\times$&8.60/6.7$\times$&4.84/7.2$\times$\\
				\noalign{\smallskip}
				\hline
			\end{tabular}
	\end{center}
	\vskip -0.2in
\end{table} 

Observing our GoogLeNet\footnote{ The learning rate 
	tuning policy of GoogLeNet used was: $$\eta = \eta_0 (1-\textsc{epoch}{}\times\frac{\textsc{number of minibatches}}{\textsc{max iterations}})^{0.5}.$$ The learning rate 
	tuning policy of AlexNet used was: scaling down by a factor of 10 every 20 epochs.} benchmark result in Fig.~\ref{f:SUBGD_googlenet}, 
we would expect that the framework provides a convergence speedup close to 
the throughput speedup reported in Table \ref{t:cross}, if the convergence 
of parallel SGD closely approximates that of sequential SGD. However, it is 
difficult to give the exact convergence speedup provided by the framework, 
since different settings of the hyper-parameters (learning rate tunning policy, 
weight decay, batch size, cropping randomness) leads to a different 
convergence path and complicates comparison.

Besides the synchronous framework, we also explored reducing the
communication overhead in the asynchronous setting. Referencing the
implementation of EASGD in \emph{Platoon}, a Theano-based multi-GPU
framework that exploits data parallelism, we re-implemented the
framework based on the CUDA-aware MPI \verb|SendRecv()| function
without the Round-Robin scheme \cite{Zhang15}. Our test shows, when training 
AlexNet on 8 GPUs, the asynchronous communication overhead in our framework is 42\% 
lower than that in Platoon when worker processes communicate with the server in the 
most frequent way ($\tau=1$). 
We also performed a grid
search on the hyper-parameters $\alpha$ and $\tau$ to achieve better
convergence when training AlexNet on eight distributed GPUs, each processing
a batch size of 128. The best top-5 error we
achieved from this framework was 21.12\% at a global epoch of 49 when the moving rate was
$\alpha=0.5$ and averaging period was $\tau=1$ with a data throughput speedup
of 6.7$\times$.

\section{Hardware and Software Environment}

\begin{figure}[h]
	\begin{center}
		\centerline{\includegraphics[scale=0.5]{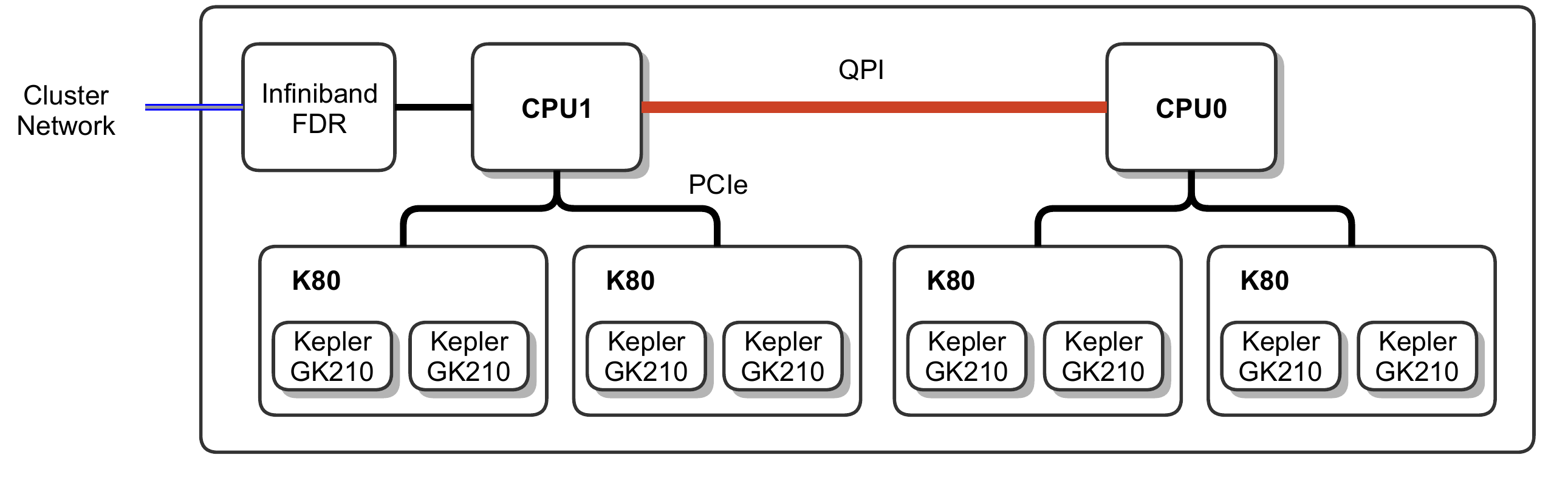}}
		\caption{Hardware connection layout of a copper node}
		\label{f:copper}
	\end{center}
	\vskip -0.3in
\end{figure}

The software was developed and tested on a
PI-contributed SHARCNET cluster, named \emph{copper}. 
As shown in Fig.~\ref{f:copper},
each node in the cluster is a dual socket system with two NVIDIA Tesla
K80 GPUs on each socket. 
The whole cluster is inter-connected with Mellonox
Infiniband FDR. We also tested on another
cluster, \emph{mosaic}, which features distributed GPUs 
across nodes connected by Infiniband QDR. 
Each node has one NVIDIA K20m GPU.

For high-level access to MPI functionality, we use its Python
binding \verb|mpi4py|, compiled against OpenMPI 1.8.7.
All models mentioned in this report are constructed in
Theano 0.8 and their implementation is available in our Github
project. Convolution and pooling operations in the
computational graph depend on CUDA 7.0 and the \verb|cuDNN| v4 library.
We also support \verb|cudaconvnet| as an alternative
backend.

\section{Discussion}

We have attempted to scale up the training of deep learning
models in an accessible way by developing a scalable training framework
built around Theano.
Key technical features of our framework are more efficient interprocess communication
strategies and parallel data loading techniques. Factors affecting the speedup
of the framework can be associated with the model to be trained (i.e.~architectural), 
the training data loading strategy, synchronization in the
computational graph, implementation of GPU kernels, system memory
and network bandwidth.

Importantly, we try not to compromise the convergence of models trained
under our framework since measured speedup
is based on the time taken to reach a certain error
rate. However, the convergence achieved by a parallel framework
also depends on the tuning of that framework's hyper-parameters.
The convergence results in Table \ref{t:lr} can therefore
be improved if better hyper-parameters are found. Factors
affecting model convergence include the number of worker processes,
effective batch size and corresponding learning rate,
parameter averaging frequency $\tau$\footnote{In BSP,
	we use $\tau=1$ since larger $\tau$ tends to affect
	convergence in the same way as increasing batch
	size.}, moving rate $\alpha$ in EASGD and the
initialization of model parameters.

The main contributions of our work include: providing multi-node and 
improved multi-GPU support to the Theano library based on MPI, eliminating 
substantial communication overhead, exposing convergence and speedup 
phenomena in parallel SGD, and an implementation of a more efficient parallel 
loading method. 

Our effort towards eliminating the communication overhead involves
several aspects: leveraging CUDA-aware MPI for direct data transfer,
separating data transfer and summation for more efficient summation on
GPUs, and exploring half precision data transfer for faster
communication.  Our benchmarking results show that our effort on
eliminating communication overhead works well on both the 1-GPU-per-node
cluster, \emph{mosaic}, and the 8-GPU-per-node cluster,
\emph{copper}. 

Note that the multi-node testing results in this report are obtained
\emph{without} GPUDirect RDMA support due to a limitation in the
cluster configuration. Also, the QPI bus topology of a \emph{copper}
node limits the usage of GPUDirect P2P technology. This is because the
GPUDirect P2P requires all GPUs to be under the same PCIe switch. If a
path traversing the QPI is needed, the data transfer would go through
CPU RAM first. As a result, further improvement of communication
performance based on the current hardware setting would involve
consideration of overlapping data transfer with the summation kernel, 
overlapping parameter exchange with gradient calculation, and designing better 
inter-node and intra-node strategies 
that could balance the bandwidth usage among QPI, PCIe and Infiniband networking.


\bibliographystyle{splncs03}
\bibliography{paper}

\end{document}